\documentclass[letterpaper, 10 pt, conference]{ieeeconf}  % Comment this line out if you need a4paper

\usepackage{amsmath,amsfonts}
\usepackage{algorithmic}
\usepackage{algorithm}
\usepackage{array}
\usepackage[caption=false,font=normalsize,labelfont=sf,textfont=sf]{subfig}
\usepackage{textcomp}
\usepackage{stfloats}
\usepackage{url}
\usepackage{verbatim}
\usepackage{graphicx}
\usepackage{cite}
\usepackage{booktabs}
\usepackage{makecell}
\usepackage{multirow}
\usepackage{float}
\usepackage{bbding}
\usepackage{wrapfig,lipsum}
\usepackage{subfig}
\usepackage{mathtools}

\IEEEoverridecommandlockouts                              % This command is only needed if 
\title{\LARGE \bf
EFFOcc: Learning Efficient Occupancy Networks from Minimal Labels for Autonomous Driving
}

\begin{document}
\pagestyle{empty}  % no page number for the second and the later pages
\thispagestyle{empty} % no page number for the first page

\author{Yining Shi$^{1,2}$, Kun Jiang$^{1,2\dagger}$, Jinyu Miao$^{1,2}$, Ke Wang$^{3}$, Kangan Qian$^{1,2}$, Yunlong Wang$^{1,2}$, Jiusi Li$^{1,2}$, \\Tuopu Wen$^{1,2}$, Mengmeng Yang$^{1,2}$, Yiliang Xu$^{4}$, Diange Yang$^{1,2\dagger}$
% <-this % stops a space
\thanks{$^{1}$ School of Vehicle and Mobility, Tsinghua University, $^{2}$ State Key Laboratory of Intelligent Green Vehicle and Mobility, Beijing, China. $^{3}$ Kargobot. Inc, $^{4}$ Zongmu Technology. This work was done during Yining Shi's internship at Kargobot. $\dagger$ Corresponding authors: Diange Yang, Kun Jiang (ydg@mail.tsinghua.edu.cn, jiangkun@mail.tsinghua.edu.cn.)}
}% <-this % stops a space
% % \thanks{Manuscript received April 19, 2021; revised August 16, 2021.}}

% \author{Albert Author$^{1}$ and Bernard D. Researcher$^{2}$% <-this % stops a space
% \thanks{*This work was not supported by any organization}% <-this % stops a space
% \thanks{$^{1}$Albert Author is with Faculty of Electrical Engineering, Mathematics and Computer Science,
%         University of Twente, 7500 AE Enschede, The Netherlands
%         {\tt\small albert.author@papercept.net}}%
% \thanks{$^{2}$Bernard D. Researcheris with the Department of Electrical Engineering, Wright State University,
%         Dayton, OH 45435, USA
%         {\tt\small b.d.researcher@ieee.org}}%
% }

\maketitle
\thispagestyle{empty}
\pagestyle{empty}

\begin{abstract}
3D occupancy prediction (3DOcc) is a rapidly rising and challenging perception task in the field of autonomous driving. Existing 3D occupancy networks (OccNets) are both computationally heavy and label-hungry. In terms of model complexity, OccNets are commonly composed of heavy Conv3D modules or transformers at the voxel level. Moreover, OccNets are supervised with expensive large-scale dense voxel labels. Model and data inefficiencies, caused by excessive network parameters and label annotation requirements, severely hinder the onboard deployment of OccNets. This paper proposes an EFFicient Occupancy learning framework, EFFOcc, that targets minimal network complexity and label requirements while achieving state-of-the-art accuracy. We first propose an efficient fusion-based OccNet that only uses simple 2D operators and improves accuracy to the state-of-the-art on three large-scale benchmarks: Occ3D-nuScenes, Occ3D-Waymo, and OpenOccupancy-nuScenes. On the Occ3D-nuScenes benchmark, the fusion-based model with ResNet-18 as the image backbone has 21.35M parameters and achieves 51.49 in terms of mean Intersection over Union (mIoU). Furthermore, we propose a multi-stage occupancy-oriented distillation to efficiently transfer knowledge to vision-only OccNet. Extensive experiments on occupancy benchmarks show state-of-the-art precision for both fusion-based and vision-based OccNets. For the demonstration of learning with limited labels, we achieve 94.38\% of the performance (mIoU = 28.38) of a 100\% labeled vision OccNet (mIoU = 30.07) using the same OccNet trained with only 40\% labeled sequences and distillation from the fusion-based OccNet. Code will be available at \url{https://github.com/synsin0/EFFOcc}.

\end{abstract}

% \begin{IEEEkeywords}
% Autonomous driving, 3D occupancy prediction, Multi-sensor fusion, Knowledge distillation
% \end{IEEEkeywords}

\section{Introduction}
Autonomous perception requires a comprehensive understanding of the environment. Common object-centric pipelines, which consist of detection, tracking, and prediction, represent obstacles as bounding boxes. It is difficult for these pipelines to deal with extra-long, irregularly shaped objects. In recent years, there has been a revival of occupancy grids in autonomous perception \cite{grid_centric_survey}. Tesla pioneers the extension from an occupancy grid map (OGM) to an occupancy network (OccNet). Tesla's vision-based Occupancy Network \cite{tesla_wad22_cvpr}  uses deep learning techniques to project visual features into 3D voxels and decode a variety of information such as occupancy, semantics, and motion flow. 

\begin{figure}
    \centering
    \includegraphics[width=0.45\textwidth]{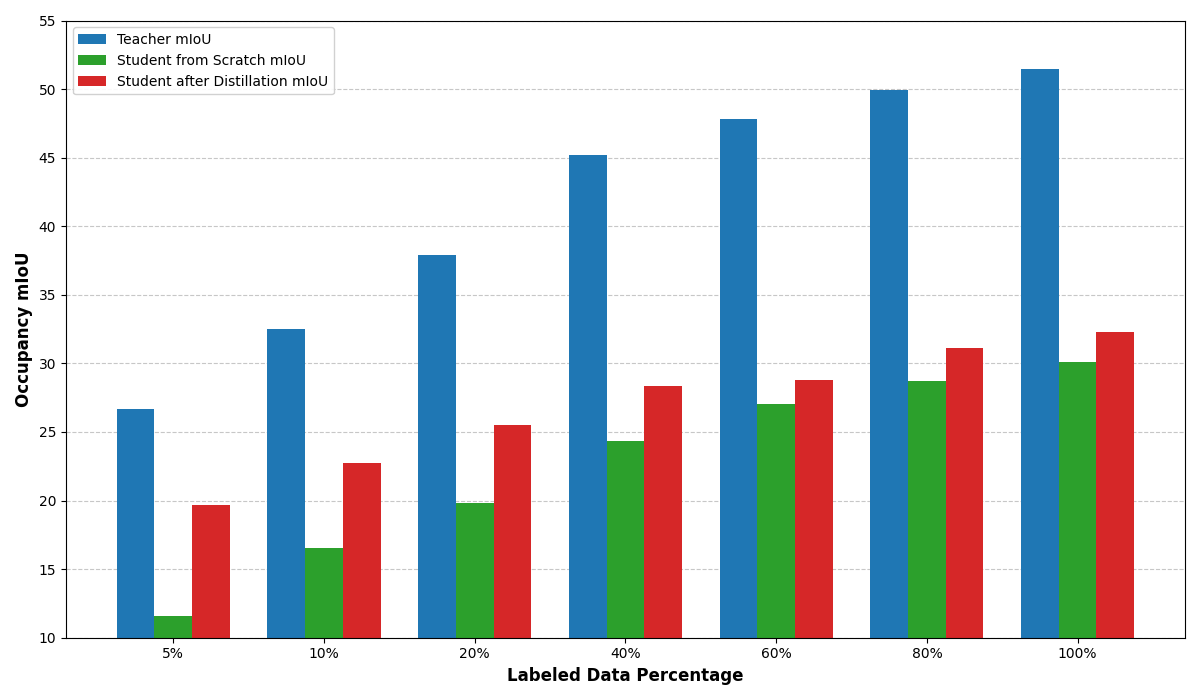}
    \caption{Graphical statistics of fusion-based teacher models and vision-only OccNets trained from scratch and trained with distillation under different labeled data scales. }
    \vspace{-3mm}
    \label{fig:effocc_distill_chart}
\end{figure}

Despite the recent success of vision-based occupancy networks, vision-based OccNets typically require a large amount of annotated data to be fully trained to a high degree of accuracy. This observation motivates us to explore a label-efficient approach to train high-accuracy vision OccNets. Since occupancy annotation is a secondary annotation based on point cloud segmentation and object detection annotation, we generally automatically annotate the entire continuous scene. Therefore, we need to use a scene as a minimum unit for efficient learning. We hope to train OccNets with the minimum number of scenes to be as accurate as possible. We find that LiDAR-camera fusion occupancy networks can better fit a small amount of annotated data. We propose to distill occupancy knowledge from a fusion-based teacher model to a vision-based student model with both labeled and unlabeled data.

% Following this trend, new 3D occupancy benchmarks\cite{Occ3D,OpenOccupancy} are built upon large-scale public datasets. These benchmarks formulate the task as semantic segmentation of foreground objects and background stuff on 3D voxel grids. 

% \iffalse
% For example, extra-long objects require the object detection network to have a large receptive field. Given unstable and inaccurate detection of general objects, autonomous vehicles require the ability to stably perceive and avoid obstacles of any type within a limited range, while taking into account end-to-end delay and balance mis-detection.
% \fi

% Despite recent success of high performance OccNets, they bring a large amount of computational load which refrains real-time onboard deployment. Reduction of model complexity is one of the major directions of improving OccNets\cite{FlashOcc,FastOcc, SparseOcc, SparseOcc_noah}. However, existing computationally-efficient OccNets focus on vision networks, while lightweight LiDAR-camera fusion OccNets are rarely explored.

To this end, we introduce EFFOcc, a novel occupancy learning framework towards learning efficient and high-performance 3D occupancy networks from minimal labels.

We propose a computationally efficient fusion-based OccNet which achieves state-of-the-art occupancy prediction performance with fewer parameters. Our motivation starts from the fact that the LiDAR point cloud is naturally suitable for geometry reconstruction while a lightweight vision branch is readily enough to compensate for semantic recognition capability. We design a lightweight fusion network and discuss multiple training techniques to achieve state-of-the-art performance. 

% The comparison between EFFOcc and other fusion OccNet is shown in Fig. \ref{fig:compare_effocc}.

% \begin{figure}
%     \centering
%     \includegraphics[width=0.5\textwidth]{fig/compare.png}
%     \caption{Comparison of two variants of EFFOcc with other state-of-the-art OccNets on Occ3D-nuScenes and OpenOccupancy-nuScenes benchmarks. EFFOcc-T uses ResNet18 as image backbone while EFFOcc-B and RadOcc-LC uses Swin-B as image backbone. We achieve better accuracy(top left), less parameters (top right), faster speed (bottom left), than RadOcc-LC. On OpenOccupancy-nuScenes dataset, EFFOcc-T model beats other fusion OccNets in terms of geometric IoU and semantic mean IoU.}
%     \vspace{-3mm}
%     \label{fig:compare_effocc}
% \end{figure}

% We conduct efficient active learning to figure out what is the minimal data requirement of fusion-based OccNet training. As the label generation process requires a lot of pre-processing, such as aggregation, matching, occlusion elimination, we believe that reducing the need of number of labelled voxels can significantly reduce the cost. We propose an entropy-based active learning method for label-efficient adaptive annotation. 

% The first stage selects a certain proportion of high-value frames from the unlabeled pool as candidates, and the second stage selects high-value voxels from high-value frames as the final selections for active learning.

We distill occupancy knowledge from the fusion-based teacher model to the vision-based student model with a small portion of labeled data and other unlabeled data. We believe that knowledge distillation helps to improve the performance of vision-only OccNets beyond occupancy label supervision, especially for unlabeled data. In the knowledge distillation process, we first decompose the occupancy result of the teacher model into three sub-regions: foreground, background, and empty regions. Then the region masks are both applicable to BEV and 3D regions. We conduct both BEV-region and 3D-region feature distillation from the teacher model to the vision-only student model. The benefits of distillation at different data scales are shown in Fig. \ref{fig:effocc_distill_chart}.

In summary, our contributions are listed as follows:

\begin{itemize}
\item We propose a label-efficient occupancy learning framework, EFFOcc, that effectively and efficiently trains fusion-based and vision-based OccNets. We provide a simple fusion-based OccNet for LiDAR-camera fusion-based 3D occupancy prediction and discuss training techniques to lift the fusion model to the state-of-the-art with lightweight designs. 

\item We propose a multi-stage occupancy-oriented distillation method to distill a real-time vision-only occupancy network with the fusion-based teacher model and get competitive performance with fewer labels compared to other real-time vision OccNets.

\item We validate our models on three public benchmarks on two large-scale datasets, nuScenes and Waymo Open Dataset, and demonstrate their effectiveness.

\end{itemize}
\section{Related Works}

\subsection{Computationally-efficient Occupancy Networks} 

\begin{figure*}[ht]
	\centering
\includegraphics[width=0.8\textwidth]{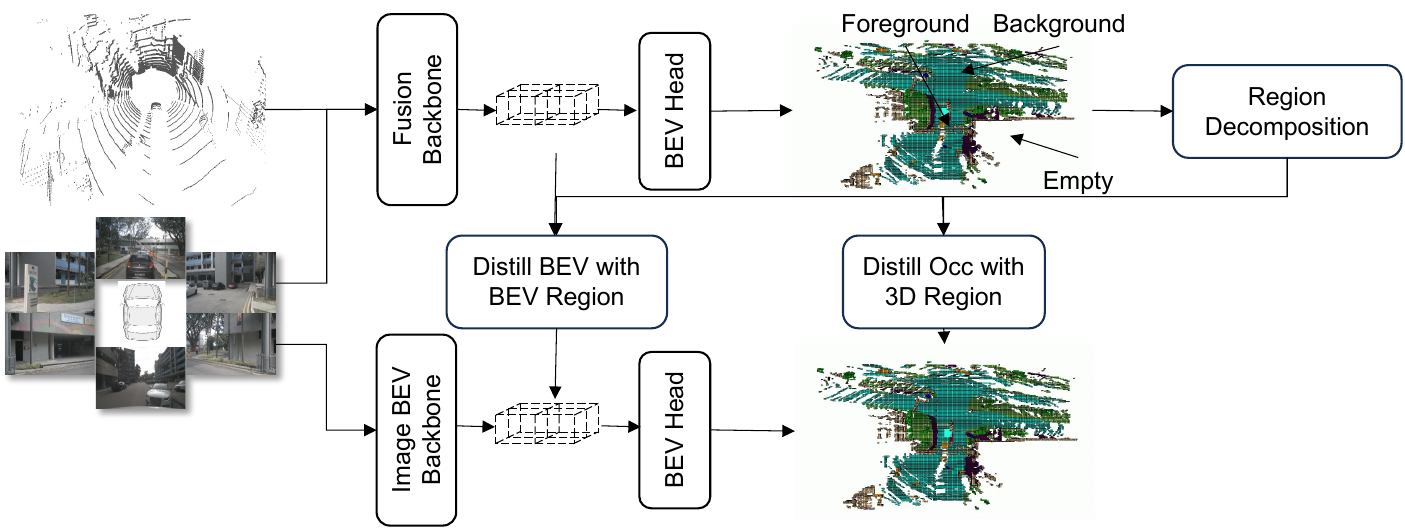}
	%\vspace{-7mm}
	\caption{The framework of EFFOcc. The LiDAR point cloud and multi-view images go through a fusion network for fusion-based occupancy prediction as the teacher model. The student model inputs multi-view images and distills multi-stage features from the teacher model in both BEV and 3D occupancy feature space. 
	}
	\label{fig:arch}
	%\vspace{-6mm}
\end{figure*}

3D occupancy networks (OccNets) describe the world with uniformly partitioned 3D voxels. OccNets can accurately reconstruct general objects under occlusion, but they usually bring a huge amount of computation load for on-board deployment. Different methods are proposed in an effort to reduce the amount of 3D voxel calculations. PanoOcc\cite{PanoOcc} replaces 3D Conv operators with sparse conv at each layer while predicting the occupancy rate of nonempty voxels and deleting predicted empty voxels to maintain sparsity. FlashOcc\cite{FlashOcc} and FastOcc\cite{FastOcc} propose efficient channel-to-height devoid of complex 3D convolution computation. SparseOcc from Nanjing University (NJU)\cite{SparseOcc} proposes fully sparse model to exploit geometry sparsity and sparse instance queries to fit object sparsity with mask transformers. SparseOcc from Shanghai Jiao Tong Univeristy (SJTU)\cite{SparseOcc2_noah} removes empty voxels after the geometry-based view transformation and uses spconv operators after that. Moreover, a sparse latent diffuser is proposed to diffuse empty voxels adjacent to occupied voxels. They achieve a remarkable 74.9\% reduction of FLOPs. Most computationally efficient designs are designed for vision-only OccNets, while EFFOcc first explores efficient fusion networks that perform well when trained with limited labels.

% \subsection{Active Learning for Autonomous Perception}
% Active learning is a label-efficient approach which effectively enhance training performance and reduces label costs is active learning, or dataset distillation, which is already explored in LiDAR-based 3D detection\cite{crb_3D_det} and LiDAR segmentation\cite{LESS, Annotator}.  Annotator\cite{Annotator} proposes a voxel-centric active learning paradigm which actively labels points insides certain voxels where voxel confusion degree (VCD) is high. They demonstrate competitive performance with model trained with actively selected labels with a portion of less than 1\% labelled minimum unit. Considering the labelling process of 3D occupancy, we propose a two-stage active learning strategy which selects both high value samples and high value samples in selected sequences.

\subsection{Knowledge distillation for Autonomous Perception}

Knowledge distillation is a learning technique that aims to transfer knowledge from a large complex teacher model to a small student model. Knowledge distillation is widely applied on vision BEV detection learning from teacher models (e.g. LiDAR-based or fusion-based detectors) \cite{UniDistill, distillBEV}. However, the occupancy task is more challenging compared to the detection task as occupancy prediction has a more severe class imbalance, not only class imbalance between foreground objects but also background stuff elements. As a result, current vision-based occupancy networks suffer from low accuracy for foreground obstacles. To distill between voxel features, RadOcc\cite{RadOcc} applies neural rendering to the image plane as auxiliary supervision for distillation on voxel features. Compared to prior arts, EFFOcc sets a semi-supervised knowledge distillation problem: Given limited labels and enough data, EFFOcc proposes to distill teacher model features from both labeled and unlabeled samples.

\section{Methodology}

\subsection{Task Formulation of 3D Occupancy Prediction}

The occupancy prediction task is formulated as building a semantic 3D occupancy grid with a fixed perception range and resolution. The sensor information is collected from surround-view cameras and 360-view LiDAR. Each voxel grid is represented by semantic categories.

\subsection{Architecture}
The framework of EFFOcc is shown in Fig. \ref{fig:arch}. The overall goal is to pursue the minimization of the network and the lowest cost of training from a model-centric perspective. 
Section \ref{sec:effocc_fusion} introduces the lightweight fusion-based network design with only 2D operators. 
Section \ref{sec:effocc_distill} uses the fusion-based OccNet as the teacher model and occupancy results from the teacher model as regions of interest. The proposed region-decomposed distillation improves the performance of vision-only OccNet with different portions of labeled data. 

% Section \ref{sec:effocc_al} proposes the two-stage active leaning based on maximum entropy frame and voxel selection to explore extreme compression of labels without a significant drop in occupancy accuracy.  

\subsection{Efficient Fusion Network}\label{sec:effocc_fusion}

\begin{figure}[ht]
	\centering
\includegraphics[width=0.5\textwidth]{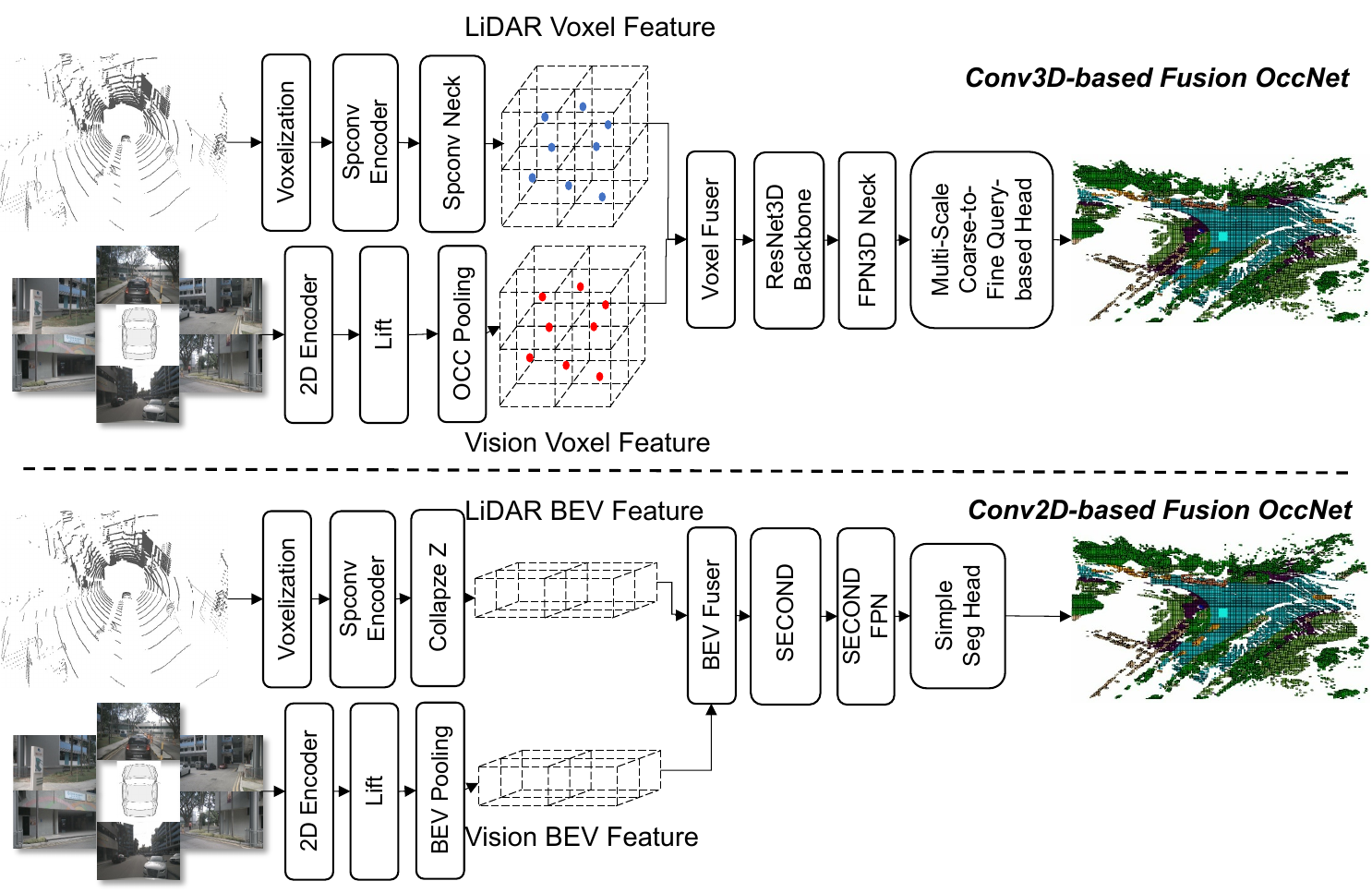}
	%\vspace{-7mm}
	\caption{Network details of the EFFOcc fusion-based OccNet framework compared to dense fusion OccNets\cite{OpenOccupancy, RadOcc}. Our lightweight design replaces the voxel features with BEV features, OCC pooling with BEV pooling, the ResNet3D backbone with the SECOND backbone, and the complex coarse-to-fine prediction head with a simple Conv2D head.
	}
	\label{fig:occ_framework_compare}
	%\vspace{-6mm}
\end{figure}

Our design goal is to achieve similar accuracy with the minimal possible network parameters.  We start from the voxel-level dense fusion method introduced in OpenOccupancy\cite{OpenOccupancy}. It consists of a visual branch, a point cloud branch, an adaptive voxel fusion module supported by 3D convolution operators, and a multi-scale segmentation head with coarse-to-fine query. We replace each module with a lightweight version without losing performance accuracy. We remove 3D CNN on the OpenOccupancy LiDAR branch, replace all Occ pooling and Conv3D Occ encoder with BEVpoolv2\cite{DAL} and Conv2D BEV encoder on the OpenOccupancy vision branch, and replace the voxel fusion layer with BEV fusion layer. Moreover, the model only uses a single-scale feature map and single-stage coarse prediction. The comparison between our network and voxel-level dense fusion method is shown in Fig. \ref{fig:occ_framework_compare}.

For pointcloud branch, we use mean feature encoding as voxel feature encoding (VFE) layer and Spconv8x with downsample stride 8 as the LiDAR encoder. Then the sparse 3D features are splatted to BEV features. For image branches, We use a image encoder and adopt BEVpoolv2\cite{DAL} as view projector to accelerate transformation from perspective view to BEV. We adopt simple conv2d-based operator as the fusion layer. After the fusion layer, We use the lightweight SECOND\cite{second} and SECOND FPN as the BEV to benefit more from detection pre-training. The occupancy head consists of two Conv2d layers and the height channel is detached from the feature channel for the final 3D output.

% \textbf{Training strategies.} We initialize from checkpoints of DAL\cite{DAL} instead of random initialization cause we find the detection pretraining improve foreground segmentation accuracy. 

We use losses from previous work. They are cross-entropy loss $\mathcal{L}_{\rm{ce}}$, Lovász-softmax loss $\mathcal{L}_{\rm{ls}}$\cite{lovasz}, affinity loss $\mathcal{L}_{\rm{scal}}^{\rm{geo}}$ and $\mathcal{L}_{\rm{scal}}^{\rm{sem}}$\cite{MonoScene}. For the Occ3D-Waymo case, we use Online Hard Example Mining (OHEM)\cite{OHEMLoss} loss. The Lovász-softmax loss and affinity loss consume more GPU memory, and improve greatly on OpenOccupancy benchmark, but help less ($<1.0$ in terms of mIoU) on Occ3D benchmark. For most experiments on the Occ3D-nuScenes benchmark, unless especially mentioned, we only use cross-entropy loss to save GPU memory during the training stage. The total loss $\mathcal{L}_{\rm{total}}$ is the weighted sum of each loss,
\begin{equation}
    \mathcal{L}_{\rm{total}} = w_{ce} \cdot \mathcal{L}_{\rm{ce}} + w_{ls} \cdot \mathcal{L}_{\rm{ls}} + w_{geo} \cdot\mathcal{L}_{\rm{scal}}^{\rm{geo}} + w_{sem} \cdot\mathcal{L}_{\rm{scal}}^{\rm{sem}}
\end{equation}
We set all weights $w_{ce}=w_{ls}=w_{geo}=w_{sem}=1$.

\subsection{Multi-stage Occupancy Distillation}\label{sec:effocc_distill}
The distillation process transfers the knowledge of the fusion-based OccNet to improve the vision-based OccNet. 
To start with, we first conduct a naive distillation which simply performs full-space feature alignment between BEV features generated from the fusion-based teacher model and vision-based student model, but fails to improve accuracy. One possible reason is that the occupancy network needs to deal with the foreground, background, and empty surroundings at the same time, and faces a severer unbalanced semantic distribution. Our statistics on BEV feature maps finds that less than 1\% pillars are with foreground objects, around 40\% pillars are with background, while the rest pillars are all empty. We design the distillation strategy to focus more on foreground voxels. 

Inspired by multi-stage distillation practices of BEV detectors, we perform distillation on both BEV space and 3D space. We decompose the foreground, background and empty region of teacher model prediction and feed regions into both BEV space and 3D space.  
The distillation loss between BEV feature $L_{\text{d\_bev}}$ and between 3D feature $L_{\text{d\_occ}}$ from teacher $F_t$ and student $F^s$ are: 
\begin{equation}
L_{\text{d\_bev}} = \sum_{i}^{\{\text{f}, \text{b}, \text{e}\}} \frac{w_i}{|S_i|} \sum_{x}^W \sum_{y}^L S_i^{(x, y)} \times \left( F_t^{(x, y)} - F^{(x, y)} \right)^2
\end{equation}

\begin{equation}
\begin{split}
L_{\text{d\_occ}} = \sum_{i}^{\{\text{f}, \text{b}, \text{e}\}} \frac{w_i}{|S_i|} \sum_{x}^W \sum_{y}^L \sum_{z}^H S_i^{(x, y, z)} \\
\times \left( 1- \cos(F_t^{(x, y, z)} - F^{(x, y, z)}) \right)
\end{split}
\end{equation}

\begin{equation}
  L_{\text{distill}} = w_{\text{bev}} \cdot L_{\text{d\_bev}} + w_{\text{occ}} \cdot L_{\text{d\_occ}}
\end{equation}

where foreground region mask $S_f$, background region mask $S_b$, and empty region mask $S_e$ has region weights $w_f$, $w_b$, $w_e$. $x, y, z$ denote the coordinate indexes on a BEV or 3D feature map. All region masks are binary. $W, L, H$ is the width, length and height of the 3D feature map. $|S_f|$ is the sum of foreground region grids. We empirically set $w_f=w_b=w_e=1$ for an equal average of three regions. We compute mean-square error (MSE) loss for BEV feature map and cosine similarity loss for 3D feature map. The final distillation loss $L_{\text{distill}}$ is the weighted ($w_{\text{bev}}$ and $w_{\text{occ}}$) sum of distillation losses on both BEV $L_{\text{d\_bev}}$ and 3D space $L_{\text{d\_occ}}$.

The vision-only student network is trained with the sum of distillation loss and classification loss when the input data is labeled and distillation loss only when the input data is unlabeled.

\section{Experiments}

\begin{table}[h] %{0.45\linewidth}
    \centering
    \caption{
     Efficient distillation results with limited labeled data on Occ3D-nuScenes benchmarks. 
    }
    \resizebox{0.5\textwidth}{!}{
    \begin{tabular}{lll|l}
    \toprule
    Data & \makecell{Fusion mIoU \\ Teacher Model} & \makecell{Vision mIoU \\ From Scratch } &  \makecell{Vision mIoU \\ After Distillation } \\\midrule
    5\% & 26.65 & 11.59 & 19.66 \\
    10\% &  32.49 & 16.55 &  22.71   \\
    20\% &  37.90  & 19.79 &  25.52   \\
    40\% &  45.19  & 24.31 &  28.38   \\
    60\% &  47.81  & 27.02 &  28.80   \\
    80\% &  49.90  & 28.72 &  31.11   \\
    100\% & \textbf{51.49} & \textbf{30.07} &  \textbf{32.32}   \\
    \bottomrule
    \end{tabular}
    }

    \vspace{0.1cm}
    \label{tab:efficient_learning}
\end{table}

\begin{table*}[ht]
	\footnotesize
        \caption{\textbf{3D occupancy prediction performance on the Occ3D-nuScenes validation set.} $\dagger$ denotes the performance reproduced by official codes. $^\star$~means the results provided by the original paper. We report six variants of EFFOcc fusion-based models. EFFOcc-R18$^A$ is trained with CE loss for 24 epochs. EFFOcc-R18$^B$ is trained with all four losses (cross-entropy loss, Lovász-softmax loss, geometric and semantic affinity loss) for 24 epochs. Both EFFOcc-R18$^A$ and EFFOcc-R18$^B$ are initialized from detection checkpoints of DAL\cite{DAL}. EFFOcc-R18$^C$ is trained with CE loss for 48 epochs. EFFOcc-R50 and EFFOcc-Swin-B are trained with CE loss for 24 epochs. Const. veh. refers to construction vehicles and dri. sur. refers to drivable surface.}
	\setlength{\tabcolsep}{0.0025\linewidth}
	%\vspace{-10pt}
	\newcommand{\classfreq}[1]{{~\tiny(\nuscenesfreq{#1}\%)}}  %
	\begin{center}
		\resizebox{\textwidth}{!}{
			\begin{tabular}{l|c|c|c |c c  c c c c c c c c c c c c c c c}
				\toprule
				Method & \rotatebox{90}{Modality}
				& \rotatebox{90}{Backbone} & mIoU &
				\rotatebox{90}{Others} & \rotatebox{90}{Barrier} & \rotatebox{90}{Bicycle} & \rotatebox{90}{Bus} & \rotatebox{90}{Car} &\rotatebox{90}{Const. veh} &\rotatebox{90}{Motorcycle} &\rotatebox{90}{Pedestrian} &\rotatebox{90}{Traffic cone} &\rotatebox{90}{Trailer} &\rotatebox{90}{Truck} &\rotatebox{90}{Dri. sur} &\rotatebox{90}{Other flat} &\rotatebox{90}{Sidewalk} &\rotatebox{90}{Terrain} & \rotatebox{90}{Manmade}& \rotatebox{90}{Vegetation}
				
				\\
				\midrule
				\multicolumn{21}{c}{Performances on  nuScenes Validation Set} \\
				\midrule
				% CTF-Occ & C & R101 & 28.53 & 8.09 & 39.33 & 20.56 & 38.29 & 42.24 & 16.93 & 24.52 & 22.72 & 21.05 & 22.98 & 31.11 & 53.33 & 33.84 & 37.98 & 33.23 & 20.79 & 18.00\\
				BEVFormer & C & R101 & 39.24 & 10.13 & 47.91 & 24.90 & 47.57 & 54.52 & 20.23 & 28.85 & 28.02 & 25.73 & 33.03 & 38.56 & 81.98 & 40.65 & 50.93 & 53.02 & 43.86& 37.15  \\
				PanoOcc & C & R101 & {42.13} & 11.67 & 50.48 & 29.64 & 49.44 & 55.52 & 23.29 & 33.26 & 30.55 & 30.99 & 34.43 & 42.57 & 83.31 & 44.23 & 54.40 & 56.04 & 45.94 & 40.40  \\
				BEVDet$\dagger$ & C & Swin-B & 42.02 & 12.15 & 49.63 & 25.10 & 52.02 & 54.46 & 27.87 & 27.99 & 28.94 & 27.23 & 36.43 & 42.22 & 82.31 & 43.29 & 54.62 & 57.90 & 48.61 & 43.55  \\

				RadOcc-C$^\star$ & C & Swin-B & 46.06 &9.78 &54.93 &20.44 &55.24 &59.62 &30.48 &28.94 &44.66&28.04 &45.69 &48.05 &81.41 &39.80 &52.78 &56.16 &64.45 &62.64\\
				RadOcc-LC$^\star$  & LC  & {Swin-B} & {49.38} & {10.93}& {58.23} & {25.01} & {57.89} & {62.85} & {34.04} & {33.45} & {50.07} & {32.05} & {48.87} & {52.11} & {82.9} & {42.73} & {55.27} & {58.34} & {68.64} & {66.01} \\
				\midrule
			   % EFFOcc-L(ours) & L & - & 42.56 & 9.24 & 46.49 & 8.36 & 48.86 & 56.38 & 24.08 & 21.28 & 40.69 & 25.49 & 42.04 & 43.49 & 78.54 & 37.98 & 49.85 & 55.39 & 68.62 & 66.74\\
               EFFOcc (ours) & L & - & 45.13 &	7.70& 49.55& 17.93& 55.46& 60.26& 29.11& 27.57& 51.93& 30.15& 42.12& 47.25& 77.88& 33.10& 48.88& 54.18& 68.21& 65.88 \\ 

                EFFOcc$^A$(ours) & LC & R18 & 49.29 & 10.57 & 56.16 & 21.73 & 58.68 & 63.16 & 31.98 & 37.71 & 55.4 & 36.15 & 45.87 & 50.81 & 81.02 & 39.07 & 53.08 & 57.15 & 70.41 & 68.90 \\

                % EFFOcc-4x(ours) & LC & R18 & 49.51 & 10.94& 56.36&23.31& 58.21& 63.57& 31.63& 38.14& 56.03& 37.20& 45.73& 50.80& 80.98& 39.22& 53.07& 57.14& 70.55& 68.85 \\
                % EFFOcc$^B$(ours) & LC & R18 & 49.59 & 10.76& 56.64 & 23.39& 58.45& 63.53& 32.05& 38.61& 55.93& 36.93& 45.91& 50.8& 81.08& 39.24& 53.23& 57.17& 70.29& 68.97 \\
                
                EFFOcc$^B$(ours) & LC & R18 & 50.46&14.34& 57.22& \textbf{40.82}& 57.60& 61.99& 34.93& \textbf{50.18}& 55.92& 42.9& 40.05& 50.09& 77.84& 38.6& 47.78& 54.9& 67.36& 65.31 \\

                EFFOcc$^C$(ours) & LC & R18 & 51.49&12.79& 58.94& 25.08& 58.33& 65.39& 32.85& 38.5& \textbf{57.82}& 38.91& 48.75& 51.54& 83.08& 44.46& 56.86& 60.42& \textbf{71.35}& \textbf{70.37} \\  
                % {'others': 12.79, 'barrier': 58.94, 'bicycle': 25.08, 'bus': 58.33, 'car': 65.39, 'construction_vehicle': 32.85, 'motorcycle': 38.5, 'pedestrian': 57.82, 'traffic_cone': 38.81, 'trailer': 48.75, 'truck': 51.54, 'driveable_surface': 83.08, 'other_flat': 44.46, 'sidewalk': 56.86, 'terrain': 60.42, 'manmade': 71.35, 'vegetation': 70.37, 'mIoU': 51.49}

                EFFOcc (ours) & LC & R50 & 52.82 & 12.09 & 59.67 & 33.39 & 61.76 & 64.98 & 35.46 & 46.01 & 57.09 & 41.04 & 47.87 & 54.59 & 82.76 & 43.95 & 56.37 & 60.23 & 71.12 & 69.60\\
                EFFOcc (ours) & LC & Swin-B & \textbf{54.08} & \textbf{15.74} & \textbf{60.98} & 36.21 & \textbf{62.24} & \textbf{66.42} & \textbf{38.68} & 43.88 & 52.12 & \textbf{42.40} & \textbf{50.29} & \textbf{56.08} & \textbf{84.92} & \textbf{48.00} & \textbf{58.60} & \textbf{61.99} & 71.29 & 69.48\\

				\bottomrule
			\end{tabular}
		}
	\end{center}
        
	%	\vspace{-.6cm}
	\label{tab:nusc_occ_val}%\vspace{-.3cm}
\end{table*}

\begin{table*}[ht]
	\footnotesize
        \caption{\textbf{3D occupancy prediction performance on the Occ3D-Waymo validation set.} $\dagger$ denotes the performance reproduced by official codes. $^\star$ means results provided by Occ3D\cite{Occ3D}.}
	\setlength{\tabcolsep}{0.0025\linewidth}
	%\vspace{-10pt}
	\newcommand{\classfreq}[1]{{~\tiny(\nuscenesfreq{#1}\%)}}  %
	\begin{center}
		\resizebox{\textwidth}{!}{
			\begin{tabular}{l|c|c|c |c c  c c c c c c c c c c c c c}
				\toprule
				Method & \rotatebox{90}{Modality}
				& \rotatebox{90}{Backbone} & mIoU &
				\rotatebox{90}{General object} & \rotatebox{90}{Vehicle} & \rotatebox{90}{Bicyclist} & \rotatebox{90}{Pedestrian} & \rotatebox{90}{Sign} &\rotatebox{90}{Traffic light} &\rotatebox{90}{Car} &\rotatebox{90}{Construction cone} &\rotatebox{90}{Bicycle} &\rotatebox{90}{Motocycle} &\rotatebox{90}{Building} &\rotatebox{90}{Vegetation} &\rotatebox{90}{Tree truck} &\rotatebox{90}{Road} &\rotatebox{90}{Sidewalk}
				
				\\
							\midrule
				\multicolumn{16}{c}{Training with 20\% training data for 8 epochs} \\
				\midrule
			BEVDet$^\star$ & C & R101 & 9.88 & 0.13 &13.06 &2.17 &10.15 &7.80 &5.85 &4.62 &0.94 &1.49 &0.0 &7.27& 10.06 &2.35 &48.15 &34.12  \\ 
	
		    BEVFormer$^\star$ & C & R101 & 15.62 & 2.59 &25.76 &13.87 &4.11 &14.23 &3.35 &8.41 &7.54 &3.45 &0.0 &18.46 &16.21 &6.87 &67.72 &41.68   \\
		    CTF-Occ$^\star$ & C & R101 & 18.73 & 6.26 &28.09 &14.66 &8.22 &15.44 &10.53 &11.78 &13.62 &16.45& \textbf{0.65} &18.63 &17.3 &8.29 &67.99 &42.98 \\
				\midrule
    			% EFFOcc (ours) & C & R50 & 19.20	& 5.90& 26.80& 16.41& 7.18& 12.98& 8.50& 10.84& 9.55& 4.22& 0.00& 23.16& 22.19& 7.89& 76.70& 55.64\\
                    EFFOcc (ours) & L & - & 41.62 & 4.28& \textbf{66.61}& 51.97& 34.00 & 30.82& \textbf{30.23}& 45.39& 27.16& 12.88& 0.00& \textbf{65.35}& \textbf{61.52}& 41.26& 81.93& \textbf{70.86}\\
				EFFOcc (ours) & LC  & R18 & \textbf{43.52}	& \textbf{10.04} & 65.05& \textbf{54.74}& \textbf{35.85}& \textbf{39.57}& \textbf{30.23}& \textbf{46.76}& \textbf{32.08}& \textbf{18.07}& 0.03& 62.53& 60.78& \textbf{43.41}& \textbf{83.26}& 70.42\\
                    \midrule
				\multicolumn{16}{c}{Training with 100\% training data for 24 epochs} \\
                \midrule
                 EFFOcc (ours) & LC  & R18 & 49.59 & \textbf{13.99} & 69.5 & 57.76 & 45.64& 47.50& \textbf{34.48}& 51.44& \textbf{38.49}& \textbf{40.02}& \textbf{1.65}& 69.11& 64.02& 47.52& 86.34& 76.32 \\

                 EFFOcc (ours) & L & - & \textbf{50.35} & 10.34& \textbf{75.97}& \textbf{63.90}& \textbf{46.35}& \textbf{50.14}& 33.19&\textbf{55.50}& 30.93& 27.58& 0.00& \textbf{74.10}& \textbf{72.99}& \textbf{50.11}& \textbf{86.83}& \textbf{77.37}\\
				\bottomrule
			\end{tabular}
		}
	\end{center}

	%	\vspace{-.6cm}
	\label{tab:waymo_occ}%\vspace{-.3cm}
\end{table*}

\subsection{Datasets and Metrics} 
We validate our model on three popular occupancy benchmarks: Occ3D-nuScenes\cite{Occ3D}, Occ3D-Waymo\cite{Occ3D}, OpenOccupancy-nuScenes\cite{OpenOccupancy}. The primary metric of all benchmarks for the prediction of 3D occupancy is the mean IoU from the average of all semantic categories. Let $C$ be the number of classes, where $TP_c$ , $FP_c$ and $FN_c$ correspond to the number of true positive, false positive, and false negative predictions for class $c_i$.
\begin{equation}
    mIoU=\frac{1}{C} \sum_{c=1}^{C} \frac{T P_{c}}{T P_{c}+F P_{c}+F N_{c}}
\end{equation}

\textbf{Occ3D-nuScenes} is built on the large-scale public available nuScenes dataset\cite{nuscenes}. The dataset consists of 700 training scenes, 150 scenes for validation, and 150 scenes for testing, each annotated at a keyframe rate of 2Hz. The sensor configuration of the ego vehicle is 6 ring cameras with resolution $1600\times900$ and one 32-beam LiDAR on the top roof. There are 17 categories of semantics in occ3D-nuScenes including the general object category. 

% \textbf{Occupancy flow challenge} is an extended annotation whose resolution and perception range is the same as Occ3D-nuScenes. The new labels provide voxel flow ground-truth for dynamic objects. Note that general object is not included in the occupancy flow challenge.

\textbf{Occ3D-Waymo} is built on the large-scale public-available Waymo Open Dataset\cite{waymo}.
The dataset comprises 1,000 sequences for trainval split, among which 798 sequences are allocated for the training set and the remaining 202 sequences are designated for validation. Ground-truth labels are annotated at 10Hz. The sensor configuration of the ego vehicle is 5 ring cameras with resolution $1920\times1280$ or $1920\times1080$ and five LiDARs. There are 15 categories of semantics in Occ3D-Waymo including the general object category.

For both datasets, the Occ3D splits the surrounding world into 3D voxel grids with the resolution of $[200,200,16]$. The perception range is $[-40m, -40m, -5m, 40m, 40m, 3m]$. For both datasets, the voxel size is set as $0.4m$.

\textbf{OpenOccupancy-nuScenes} is also built on nuScenes\cite{nuscenes} dataset. This benchmark is more challenging than Occ3D-nuScenes in that it requires a wider perception range $[-51.2m, -51.2m, -5m, 51.2m, 51.2m, 3m]$ and a finer resolution $[512,512,40]$ and the voxel size is set as $0.2m$. A slight difference regarding semantic categories is that OpenOccupancy ignores general object category and only has $16$ semantic categories.

\subsection{Implementation Details}

\textbf{Data Pre-processing.}
For both training and inference phrases, we first load multi-view images with camera parameters, then apply normalization, padding, and multi-scale flipping to each input image for image augmentation. We aggregate multi-sweep LiDAR point clouds and conduct random flipping on point clouds and voxel labels for BEV augmentation. We don't use any test-time augmentation techniques. We use camera mask of Occ3D benchmarks in the training stage.  

\textbf{Training and Inference.} We build our code upon the MMDetection3D version 1.0.0rc4 \cite{mmdet3D2020}. Experiments are trained on 8 2080TI GPUs or on 4 A6000 GPUs with a total batch of 16. We use AdamW optimizer with a learning rate of 0.0001 and weight decay 0.01. We use exponantial moving average (EMA) hook for better accuracy. We set batch size as 1 in the inference stage. The inference speed is measured by a single A6000 GPU.

\begin{figure}[ht]
	\centering
\includegraphics[width=0.45\textwidth]{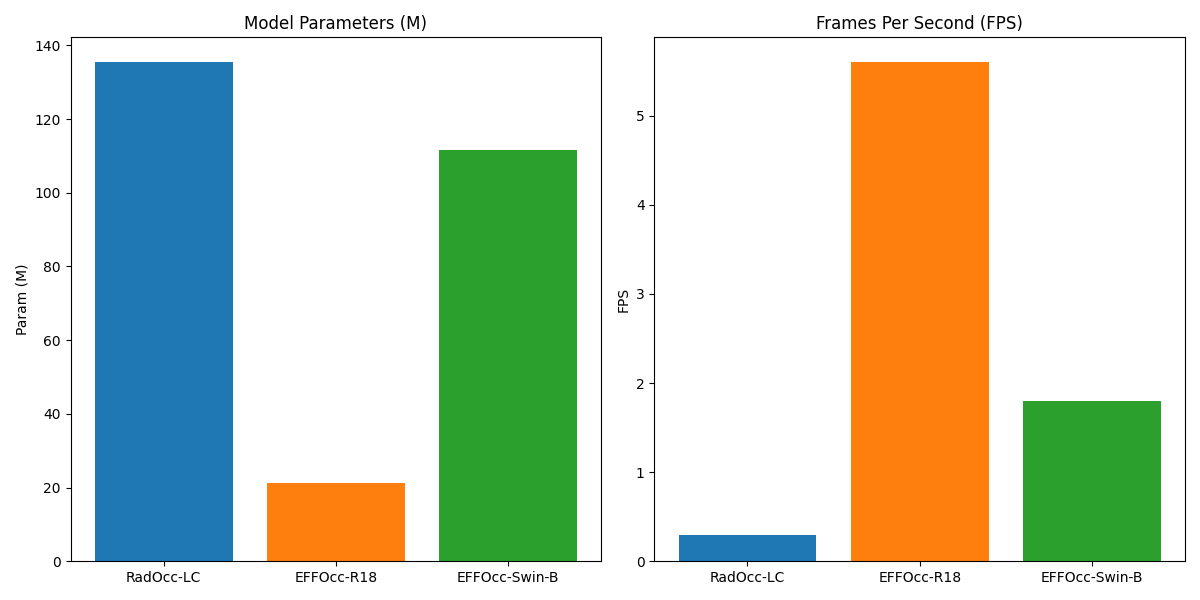}
	%\vspace{-7mm}
	\caption{Runtime efficiency analysis between RadOcc-LC, EFFOcc-R18 and EFFOcc-Swin-B. The parameters of RadOcc-LC, EFFOcc-R18, EFFOcc-Swin-B are 135.39M, 21.35M, and 111.48M. The runtime frame per second (FPS) is 0.3, 5.6, and 1.8, respectively.
	}
	\label{fig:effocc_param_fps_chart}
	%\vspace{-6mm}
\end{figure}
\subsection{Results of Efficient Learning with Limited Labels}
The overall efficient learning performance for OccNets is shown in Table. \ref{tab:efficient_learning}. We use different proportions of data to train fusion-based and vision-based networks separately, and in this setup, to illustrate the efficiency of the data, we do not use any pre-training weights for the detection network, because the detection network is trained with all the data. We can easily see that when the training data is very limited, the fusion network can achieve comparable performance with only a relatively small number of annotations, but the performance of the vision network has been low. Our distillation method is quite efficient.
With efficient learning techniques, we achieve 92.44\% of the performance (mIoU=28.38) of a 100\% labeled trained vision network (mIoU=30.07) using only 40\% of the labeled sequences. With 80\% labeled data, the distilled OccNet outperforms the 100\% labeled vision OccNet by $1.04$ mIoU.

\begin{table*}[h]
\setlength{\tabcolsep}{0.0045\linewidth}
\newcommand{\classfreq}[1]{{~\tiny(\semkitfreq{#1}\%)}}  %
\centering
\caption{\textbf{3D semantic occupancy prediction results on nuScenes-Occupancy validation set.} We report the geometric metric IoU, semantic metric mIoU, and the IoU for each semantic class. The C, and L denotes camera and LiDAR, respectively. \textbf{Bold} represents the best score. Const. veh. refers to construction vehicles and dri. sur. refers to drivable surface.}
\resizebox{1.0\linewidth}{!}{
\begin{tabular}{l|c | c c | c c c c c c c c c c c c c c c c}
    \toprule
    Method
    & \makecell[c]{Modality}
    & IoU& mIoU
    & \rotatebox{90}{Barrier}
    & \rotatebox{90}{Bicycle}
    & \rotatebox{90}{Bus}
    & \rotatebox{90}{Car}
    & \rotatebox{90}{Const. veh.}
    & \rotatebox{90}{Motorcycle}
    & \rotatebox{90}{Pedestrian}
    & \rotatebox{90}{Traffic cone}
    & \rotatebox{90}{Trailer}
    & \rotatebox{90}{Truck}
    & \rotatebox{90}{Dri. sur.}
    & \rotatebox{90}{Other flat}
    & \rotatebox{90}{Sidewalk}
    & \rotatebox{90}{Terrain}
    & \rotatebox{90}{Manmade}
    & \rotatebox{90}{Vegetation} \\
    % & mIoU\\
    \midrule
    % MonoScene~\cite{MonoScene} & C   & 18.4 & 6.9 & 7.1  & 3.9  &  9.3 &  7.2 & 5.6  & 3.0  &  5.9& 4.4& 4.9 & 4.2 & 14.9 & 6.3  & 7.9 & 7.4  & 10.0 & 7.6 \\
  
    % TPVFormer~\cite{TPVFormer} &C &  15.3 &  7.8 & 9.3  & 4.1  &  11.3 &  10.1 & 5.2  & 4.3  & 5.9 & 5.3&  6.8& 6.5 & 13.6 & 9.0  & 8.3 & 8.0  & 9.2 & 8.2 \\

    % 3DSketch~\cite{3DSketch} &  C\&D & 25.6 & 10.7  & 12.0 &  5.1 &  10.7 &  12.4 & 6.5  & 4.0  & 5.0 & 6.3&  8.0&  7.2& 21.8 &  14.8 & 13.0 &  11.8 & 12.0 & 21.2 \\
        
    % AICNet~\cite{AICNet} & C\&D   &   23.8 & 10.6  & 11.5  & 4.0  & 11.8  & 12.3&  5.1 & 3.8  & 6.2  & 6.0 & 8.2&  7.5&  24.1 & 13.0 & 12.8  & 11.5 & 11.6  &  20.2\\

    %     LMSCNet~\cite{LMSCNet} & L &   27.3 & 11.5 & 12.4&  4.2 & 12.8  & 12.1  & 6.2  &  4.7 & 6.2 & 6.3&  8.8&  7.2& 24.2 & 12.3  & 16.6 & 14.1  & 13.9 & 22.2 \\

    %     JS3C-Net~\cite{JS3C-Net} &L &   30.2  & 12.5 & 14.2 & 3.4  & 13.6  & 12.0  & 7.2  &  4.3 & 7.3 & 6.8&  9.2& 9.1 & 27.9 & 15.3  & 14.9 & 16.2  & 14.0 & 24.9 \\
         
        OccFusion~\cite{OccFusion} & LC & 31.1 & 17.0 &15.9 &15.1 &15.8 &18.2 &15.0 &17.8 &17.0 &10.4 &10.5& 15.7 &26.0 &19.4 &19.3 &18.2 &17.0 &21.2 \\
        
        M-baseline~\cite{OpenOccupancy} & LC &   29.1 & 15.1 & 14.3 & 12.0 & 15.2 & 14.9 & 13.7 &15.0 & 13.1 &9.0 &10.0 &14.5 &23.2 &17.5 &16.1 &17.2 &15.3 & 19.5  \\ 

        M-CONet~\cite{OpenOccupancy} & LC &   29.5 & 20.1 &  23.3  & 13.3& 21.2 & 24.3& \textbf{15.3}  & 15.9& 18.0 & 13.3 & 15.3 & 20.7 & 33.2 & 21.0 & 22.5  & 21.5 &19.6 & 23.2  \\

         Co-Occ~\cite{Co-Occ} & LC &30.6&21.9&26.5 &\textbf{16.8}&22.3&27.0&10.1&20.9&20.7 &14.5 &16.4 &21.6&\textbf{36.9}&23.5&\textbf{25.5} &23.7 &20.5 &23.5 \\

         OccGen~\cite{OccGen} & LC &30.3&22.0& 24.9&16.4&\textbf{22.5}&26.1 &14.0 &20.1&21.6&14.6&\textbf{17.4}&\textbf{21.9}&35.8&\textbf{24.5}&24.7&\textbf{24.0}&20.5 &23.5 \\

         \midrule
         EFFOcc (ours) & LC & \textbf{30.8} & \textbf{22.9} & \textbf{28.1} & 16.7 & 22.1 & \textbf{27.3} & 13.0 & \textbf{24.8} & \textbf{36.2} & \textbf{22.6}&  16.8& 21.6& 29.4& 13.9& 18.2& 20.6& \textbf{26.5} & \textbf{28.8} \\ 
         
    \bottomrule

\end{tabular}}

\label{tab:nusc_openoccupancy}
\end{table*}

\begin{table*}[t]
  
  \footnotesize
  \setlength{\tabcolsep}{0.0025\linewidth}
  \centering
  \caption{\textbf{3D occupancy prediction performance of real-time vision-only models on the Occ3D-nuScenes benchmark}. ``8f'' means fusing temporal information from 7+1 frames. Const. veh. refers to construction vehicles and dri. sur. refers to drivable surface.}
  \begin{tabular}{l | cc | c | c c c c c c c c c c c c c c c c c c}
      \toprule
      Method
      & \rotatebox{90}{Backbone}
      & \rotatebox{90}{Input Size}
      & \rotatebox{90}{mIoU}
      & \rotatebox{90}{Others} & \rotatebox{90}{Barrier} & \rotatebox{90}{Bicycle} & \rotatebox{90}{Bus} & \rotatebox{90}{Car} &\rotatebox{90}{Const. veh} &\rotatebox{90}{Motorcycle} &\rotatebox{90}{Pedestrian} &\rotatebox{90}{Traffic cone} &\rotatebox{90}{Trailer} &\rotatebox{90}{Truck} &\rotatebox{90}{Dri. sur} &\rotatebox{90}{Other flat} &\rotatebox{90}{Sidewalk} &\rotatebox{90}{Terrain} & \rotatebox{90}{Manmade}& \rotatebox{90}{Vegetation} \\
      \midrule
      
      MonoScene~\cite{MonoScene} & R101 & 1600$\times$900 & 6.1  & 1.8 & 7.2 & 4.3 & 4.9 & 9.4 & 5.7 & 4.0 & 3.0 & 5.9 & 4.5 & 7.2 & 14.9 & 6.3 & 7.9 & 7.4 & 1.0 & 7.7 \\
      OccFormer~\cite{OccFormer} & R101 & 1600$\times$900 & 21.9 & 5.9 & 30.3 & 12.3 & 34.4 & 39.2 & 14.4 & 16.5 & 17.2 & 9.3 & 13.9 & 26.4 & 51.0 & 31.0 & 34.7 & 22.7 & 6.8 & 7.0 \\
      BEVFormer~\cite{BEVFormer} & R101 & 1600$\times$900 & 26.9 & 5.9 & 37.8 & 17.9 & 40.4 & 42.4 & 7.4 & 23.9 & 21.8 & 21.0 & 22.4 & 30.7 & {55.4} & 28.4 & 36.0 & 28.1 & 20.0 & 17.7 \\
      CTF-Occ~\cite{Occ3D} & R101 & 1600$\times$900 & 28.5 & 8.1 & 39.3 & \textbf{20.6} & 38.3 & 42.2 & 16.9 & \textbf{24.5} & 22.7 & 21.1 & 23.0 & 31.1 & 53.3 & 33.8 & 38.0 & 33.2 & 20.8 & 18.0 \\
      TPVFormer~\cite{TPVFormer} & R101 & 1600$\times$900 & 27.8 & 7.2 & 38.9 & 13.7 & \textbf{40.8} & 45.9 & 17.2 & 20.0 & 18.9 & 14.3 & 26.7 & \textbf{34.2} & 55.7 & 35.5 & 37.6 & 30.7 & 19.4 & 16.8 \\
      \midrule
      SparseOcc\cite{SparseOcc} (1f) & R50 & 704$\times$256 & 27.0 & 8.8 & 33.2 & 17.1 & 34.4 & 41.0 & 16.1 & 19.2 & 20.8 & 21.0 & 18.4 & 27.9 & 62.4 & 31.0 & 39.2 & 35.1 & 17.5 & 16.8 \\
      SparseOcc\cite{SparseOcc} (8f) & R50 & 704$\times$256 & 30.9 & \textbf{10.6} & 39.2 & 20.2 & 32.9 & 43.3 & 19.4 & 23.8 & \textbf{23.4} & \textbf{29.3} & 21.4 & 29.3 & 67.7 & 36.3 & 44.6 & 40.9 & 22.0 & 21.9 \\
      BEVDetOcc\cite{BEVDet} (1f) & R50 & 704$\times$256 & 31.64 & 6.65& 36.97 & 8.33& 38.69& 44.46& 15.21& 13.67& 16.39& 15.27& 27.11& 31.04& 78.7& 36.45& 48.27& 51.68& 36.82& 32.09 \\

      FlashOcc\cite{FlashOcc} (1f) & R50 & 704$\times$256 & 32.08 & 6.74& 37.65& 10.26& 39.55& 44.36& 14.88& 13.4& 15.79& 15.38& 27.44& 31.73& 78.82& 37.98& 48.7& 52.5& 37.89& 32.24 \\\midrule

      EFFOcc (ours) (1f)  & R50 & 704$\times$256 & \textbf{34.30} & 8.36 & \textbf{41.56} & 13.93& 39.83& \textbf{47.19} & \textbf{20.08} & 17.26 & 19.26 & 19.18 & \textbf{29.77} & 33.75 & \textbf{79.2} & \textbf{39.55}& \textbf{48.95}& \textbf{52.95}& \textbf{38.93} & \textbf{33.40}\\

% # {'others': 8.36, 'barrier': 41.56, 'bicycle': 13.93, 'bus': 39.83, 'car': 47.19, 'construction_vehicle': 20.08, 'motorcycle': 17.26, 'pedestrian': 19.26, 'traffic_cone': 19.18, 'trailer': 29.77, 'truck': 33.75, 'driveable_surface': 79.2, 'other_flat': 39.55, 'sidewalk': 48.95, 'terrain': 52.95, 'manmade': 38.93, 'vegetation': 33.4, 'mIoU': 34.3}

  \bottomrule
  \end{tabular}

  \vspace{1mm}

  \label{tab:occ3D-nus-vision-only}
\end{table*}

\subsection{Results of Proposed Fusion-based Occupancy Network}
% For comparison on different model scales, we provide three model variants: EFFOcc-L is the LiDAR-only model with sparse conv as LiDAR backbone; EFFOcc-C is the camera-only model which is the student model distilled from the fusion-based teacher model. The model design is very much similar to FlashOcc\cite{FlashOcc}; EFFOcc and EFFOcc is the tiny and base scale version of fusion-based model, mainly differing in the scale of image backbones. 

\subsubsection{Results on Occ3D-nuScenes}
Results on Occ3D-nuScenes validation set are shown in Table. \ref{tab:nusc_occ_val}. The baselines are state-of-the-art vision models and the teacher model of RadOcc\cite{RadOcc}, RadOcc-LC. Our LiDAR-only model achieves similar mIoU compared to state-of-the-art vision-based models on top of Swin-B\cite{swin} backbone. 
\begin{figure*}[ht]
	\centering
\includegraphics[width=1.0\textwidth]{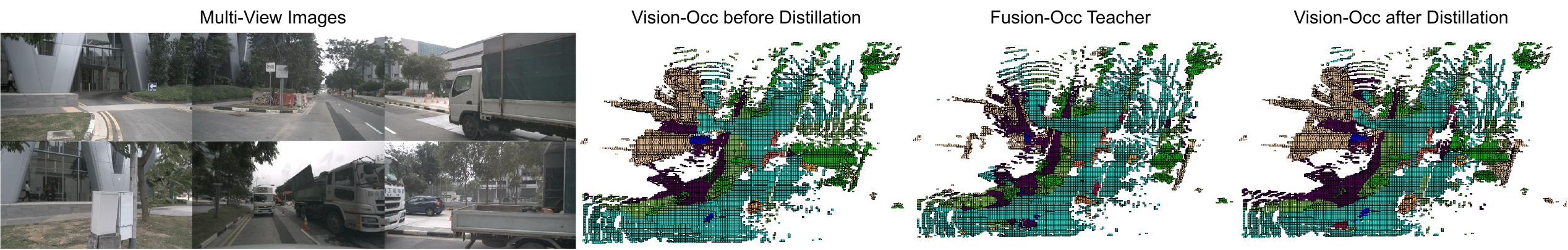}
	%\vspace{-7mm}
	\caption{Visualizations of fusion-based OccNet, vision-based OccNets before and after distillation. We use camera visibility mask in rendering of occupancy results.
	}
	\label{fig:vis_occ}
	%\vspace{-6mm}
\end{figure*}

We provide several variants of the fusion-based model to discuss training techniques to elevate fusion-based OccNet to the state-of-the-art. EFFOcc-R18$^A$, pre-trained from the detection checkpoint and using ResNet-18 as the image backbone with an image size of $256\times704$, performs only $0.09$ mIoU worse than RadOcc-LC but runs significantly faster. EFFOcc$^B$ adopts three more losses and improves $1.17$ mIoU compared to the model trained only with CE loss. But training with multiple losses doubles the training time. To this end, we try to extend the training schedule to 48 epochs and train the model from scratch, the model improves to $51.49mIoU$, which is $2.11$ mIoU higher than RadOcc-LC. Our practices demonstrate that both detection pretraining and longer training schedules help to achieve state-of-the-art performance for fusion-based OccNets.

We further scale up the image backbones and train EFFOcc-R50 and EFFOcc-Swin-B with ResNet50 and Swin-Transformer-Base as image backbones, respectively. The model performance steps higher when image backbones with more parameters are used. EFFOcc-R50 uses an image size of $256\times704$ and EFFOcc-Swin-B uses an image size of $512\times1408$. EFFOcc-Swin-B achieves a remarkable $+4.70$ mIoU gain compared with RadOcc-LC with the Swin-B backbone. 

We compare model parameters and running speed between the fusion-based baseline and our model in Fig. \ref{fig:effocc_param_fps_chart}. Our model is a highly computationally efficient occupancy network that has more than 80\% reduction in parameters and 18 times acceleration under similar mIoU precision.

\subsubsection{Results on Occ3D-Waymo}
The results on Occ3D-Waymo validation set are shown in Table. \ref{tab:waymo_occ}. To align with the evaluation of vision-based OccNets, we evaluate voxels visible to cameras. For fast validation on a smaller data scale, we follow the practice of Occ3D\cite{Occ3D} and train each model with 20\% training data for 8 epochs. LiDAR model achieves more than twice the mIoU precision as compared to vision-based methods, mainly because Waymo's LiDARs are significantly stronger than nuScenes's 32-beam LiDAR. Adding a light image branch to the LiDAR model achieves a reasonable increase of $+1.90$ mIoU. If the model is trained for 24 epochs on 100\% data with 150k training samples in total, the LiDAR-only model outperforms the fusion model by $0.76$ mIoU. The abnormal phenomenon of accuracy decrease with image branch may be due to the adequate training of the LiDAR feature network, incomplete coverage of vision, and conflict between LiDAR and vision.

\subsubsection{Results on OpenOccupancy-nuScenes} 
Results on OpenOccupancy-nuScenes validation set are shown in Table. \ref{tab:nusc_openoccupancy}. Our model is more lightweight than other models, cause we use ResNet-18 as image backbone and an image size of $256\times704$, while others use ResNet-50 and an image size of $896\times1600$. Compared with other LiDAR-camera fusion OccNets, we achieve the best semantic mIoU of $22.9$ and the best geometric IoU of $30.8$. Compared to the Occ3D-nuScenes benchmark, our method also demonstrates equally good performance under a larger perception range and finer grid resolution.

\begin{table}[h] %{0.45\linewidth}
    \caption{
     Ablation study of distillation methods, and weights of the BEV and 3D space distillation.
    }
    \tiny
    %\scriptsize
    % \footnotesize
    % \setlength{\tabcolsep}{10.0mm}
    %\small
    %\setlength{\tabcolsep}{8.0mm}
    %\normalsize
    %\large
    \centering
    \resizebox{0.5\textwidth}{!}{
    \begin{tabular}{llll|c}
    \toprule
     BEV Space & $w_{\text{bev}}$ & 3D Space & $w_{\text{occ}}$ & mIoU \\ \midrule
     - & - & - & - & 32.08 \\
     FG/BG & 1 & - & - & 33.37 \\
     MSE& 1 & - & - & 33.11 \\
     MSE& 1 & 1-cosine& 1 & \textbf{33.93} \\ 
     MSE& 1 & 1-cosine& 0.1 & 33.59 \\
     MSE& 1 & 1-cosine& 10 & 33.82 \\
     MSE& 10 & 1-cosine& 10 & 33.74 \\
     MSE& 1 & MSE& 0.1 & 33.47 \\
     MSE& 1 & CE & 1 & 31.58 \\
     MSE& 1 & KL & 1 & 31.63 \\
     MSE& 1 & FG/BG + CE& 1 & 32.77 \\
     FG/BG& 1 & 1-cosine& 1 & 33.72 \\
     FG/BG& 1 & 1-cosine& 10 & 33.75 \\
    \bottomrule
    \end{tabular}
    }

    \vspace{0.1cm}

    \label{tab:ablate_distill}
\end{table}

\subsection{Results of Vision-based OccNets with Knowledge Distillation}

The results of the vision-only occupancy network are shown in Table. \ref{tab:occ3D-nus-vision-only}. We distill the FlashOcc model which uses ResNet-50\cite{ResNet} as the image backbone and input a single-frame image with size 704$\times$256 and ResNet50 as image backbone. The model is initialized from the detection pre-trained checkpoint from BEVDet\cite{BEVDet}. We distill the vision model from the fusion-based teacher model with all labeled data for 48 epochs. In this setting, our distilled model works $2.22$ mIoU better than our baseline FlashOcc\cite{FlashOcc}.

We conduct ablation studies on different distillation methods and the results are shown in Table. \ref{tab:ablate_distill}. We start by training FlashOcc-R50 without distillation for 24 epochs. We distill on both BEV and 3D occupancy spaces. We test different loss functions, including foreground and background reweighing L1 loss (FG/BG), 1-cosine similarity (1-cosine), mean-square error (MSE), cross-entropy loss, Kullback-Leibler (KL) divergence loss. We also test different weights between BEV distillation and 3D distillation. We find that an equal weight for both BEV and 3D distillation, MSE loss for the BEV space, and 1-cosine loss for the 3D space show the best precision of mIoU=33.93.

% \subsection{Ablation Study}
% We conduct ablative experiments to demonstrate the efficacy of each component in our network design and learning strategies. Without further notice, all experiments are conducted on EFFOcc with ResNet18 as image backbone. 

% \subsubsection{Effect of Pretraining Strategy}

% We provide model details in Tab .\ref{tab:ablate_pretrain} and report their mIoUs respectively. RadOcc LiDAR and fusion-based version are our baseline with dense Conv3D operators. If model is initialized randomly, we observe an obvious gap of $3.45$ mIoU between Conv2D-based EFFOcc-L and RadOcc-L. Similarly, EFFOcc is $3.41$ mIoU inferior compared with RadOcc-LC. However, after our detection pretraining, our EFFOcc-L and EFFOcc is only $0.97$ and $0.09$ mIoU away from our baselines.

\subsection{Qualitative Results}
We visualize a one-frame example of fusion-based, vision-based and distilled occupancy predictions in Fig. \ref{fig:vis_occ}. Compared to the vision-based model before distillation, the model after distillation better aligns with the fusion-based teacher, especially in the reconstruction of distant objects. 

% \subsection{Discussions}
% Training the lightweight fusion model from scratch performs around $-3.0$ mIoU inferior with latest fusion-based RadOcc\cite{RadOcc} teacher model, which has very much similar structure with OpenOccupancy\cite{OpenOccupancy}. We find that devoid of the 3D CNN after sparse convolution(Spconv) reduces the performance. We find that one promising approach to mitigate the gap is to pretrain model parts with detection tasks. 

\section{Conclusion}
This paper mainly discusses the minimal workflow design of training occupancy networks with minimal labels and computation costs. While on par with or surpassing the performance of existing OccNets on two large-scale public datasets, we significantly reduce training costs and enhance usability. Furthermore, we design a multi-stage distillation strategy so that the fusion network can enhance the accuracy of the vision-only lightweight occupancy network. For future works, we will investigate more effective active learning techniques to search necessary minimal labels for OccNets.

\bibliographystyle{IEEEtran}
\bibliography{ref}

\end{document}